\documentclass{article}
\usepackage[preprint]{neurips_2025}

\usepackage{natbib}
\usepackage[
    colorlinks=true,
    citecolor=blue,
    linkcolor=blue,
    urlcolor=blue,
    anchorcolor=blue
]{hyperref}
\usepackage{graphicx}
\usepackage{subcaption}
\usepackage[utf8]{inputenc}
\usepackage[T1]{fontenc}
\usepackage{breakurl}
\usepackage{listings}
\usepackage{caption}
\usepackage{booktabs}
\usepackage{threeparttable}
\usepackage{placeins}
\usepackage{amsfonts}
\usepackage{amsmath}
\usepackage{amsthm}
\usepackage{xurl}
\usepackage{amssymb}
\usepackage{nicefrac}
\usepackage{microtype}
\usepackage[font=small,labelfont=bf]{caption}
\usepackage{enumitem}
\usepackage{wrapfig}
\usepackage{multirow}
\usepackage[most]{tcolorbox}
\usepackage[normalem]{ulem}
\usepackage{xspace}
\usepackage{float}
\usepackage{tabularx}
\usepackage{array}
\usepackage{adjustbox}
\usepackage{makecell}
\usepackage{pifont}
\usepackage{xcolor}

\newcommand{\cmark}{\ding{51}}%
\newcommand{\xmark}{\ding{55}}%

\title{Robostral Navigate}
\author{}

\begin{document}

\raggedbottom

\vspace{-0.1in}

\maketitle

\vspace{-55pt}

\begin{center}

    \includegraphics[
        width=0.4\linewidth,
        keepaspectratio
    ]{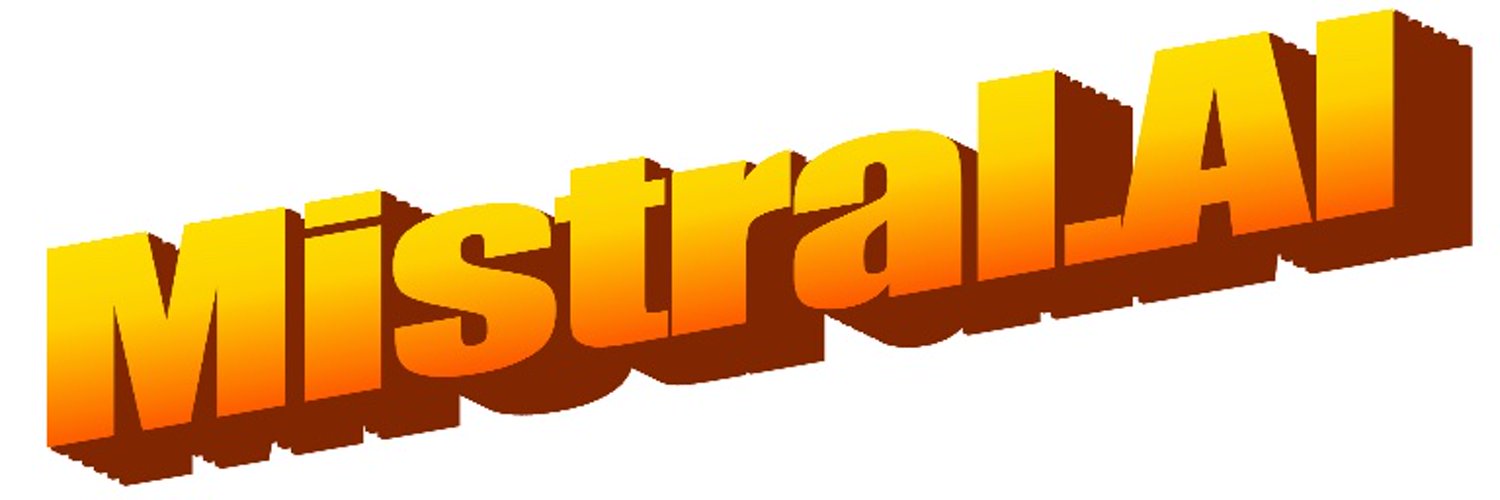}
    \vspace{2pt}

    \includegraphics[
        width=\textwidth,
        height=0.79\textheight,
        keepaspectratio
    ]{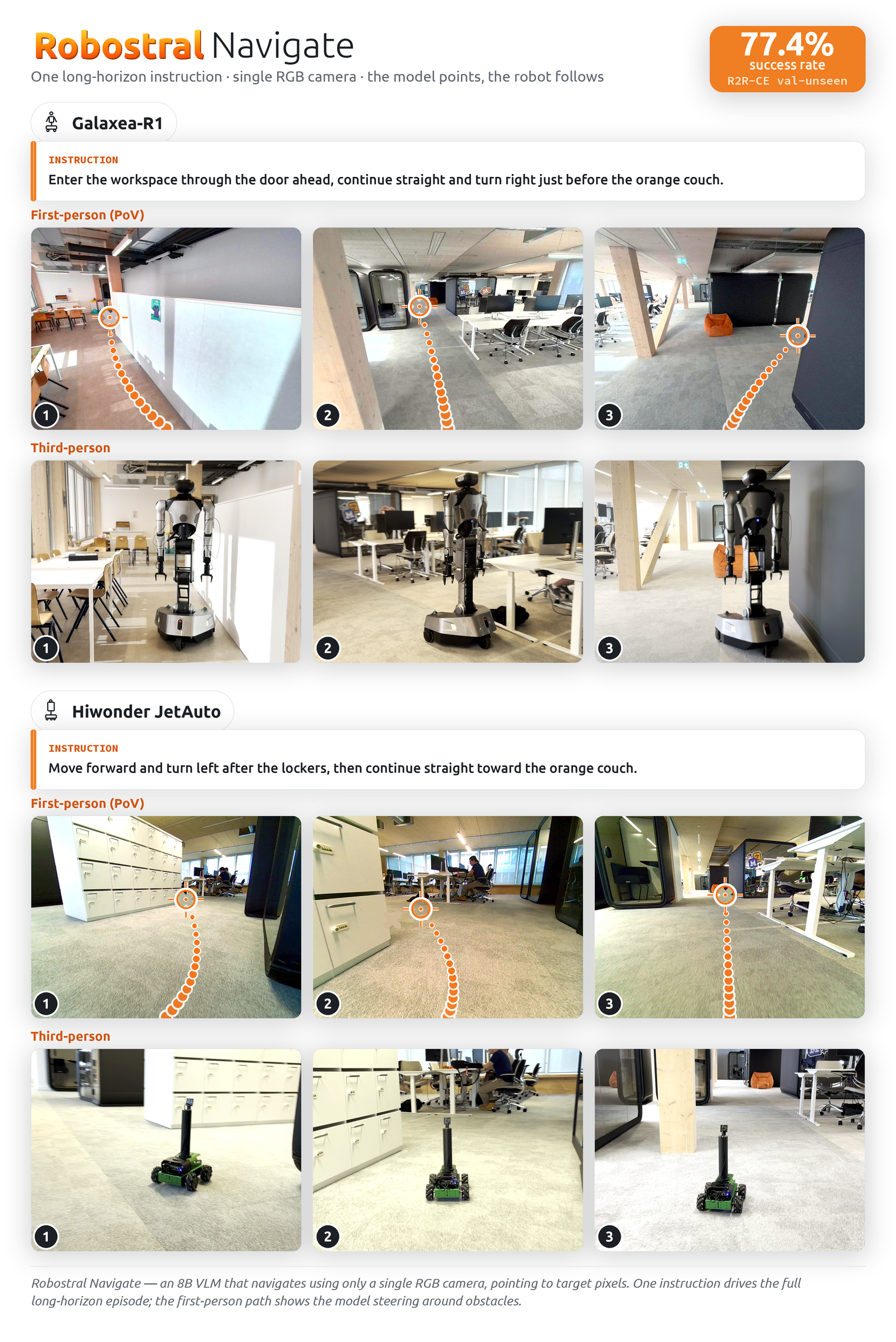}

\end{center}

\clearpage


\begin{abstract}
Deploying navigation systems at scale requires a recipe that minimizes sensor assumptions, generalizes across robot embodiments, and trains efficiently. Yet, today's best systems depend on depth sensors, multi-camera rigs, or pre-built maps, limiting the hardware they support and increasing deployment cost. We introduce Robostral Navigate, an 8B vision-language model built around this scalability objective. The model consumes only a stream of monocular RGB images—the most ubiquitous sensor across robotic platforms—and predicts waypoints by pointing to the next target location in the current camera view. Operating purely in image space, rather than robot-specific coordinates, makes the policy naturally robust to changes in camera intrinsics and scene scale, enabling deployment across wheeled, legged, and aerial robots without recalibration. We generate 2.4 million trajectories across 350k simulated scenes to reduce the reliance on real-world data collection and scale easily. We further introduce a prefix-caching training recipe that packs entire episodes into single training sequences, reducing training tokens by 22× and cutting training time from months to days. A tree-based attention mask prevents conditioning on previous ground-truth actions, encouraging visually grounded action prediction, and reinforcement learning is used to further improve exploration and recovery capabilities. On the Room-to-Room and Room-Across-Room in Continuous Environments (R2R-CE and RxR-CE) benchmarks, Robostral Navigate sets a new state of the art. On R2R-CE, it achieves a 77.4\% success rate, surpassing the best monocular method by 10.5 points and the strongest depth- or multi-camera system by 5.3 points despite using only a single RGB camera. On RxR-CE, it reaches 75.1\% success rate, outperforming all monocular baselines.

\end{abstract}

\noindent\makebox[\textwidth][c]{%
    {\small\textbf{Webpage:}}\hspace{0.3em}%
    {\scriptsize\url{https://mistral.ai/news/robostral-navigate}}%
}
\par\vspace{-0.5\baselineskip}


\section{Introduction}
Embodied navigation---the ability of a robot to move through an environment following natural language instructions---is a foundational capability for general-purpose robotics. Recent vision-language models have made impressive strides on this task ~\citep{navid, uni_navid, navila, streamvln, qwen_vla}, but the best-performing systems depend on depth sensors, LiDAR, multi-camera rigs, or pre-built environment maps~\citep{internvla_n1, navfom, omninav, abot_n0, qwen_robotnav}. Each additional sensing requirement narrows the set of compatible robots, increases per-unit cost, and demands environment-specific calibration---all of which limit how broadly a navigation policy can be deployed. As the field moves toward general-purpose robots that must operate across diverse hardware and novel environments ~\citep{brohan2023rt2, shah2023lm}, the challenge is not just building a more accurate navigator, but finding a \emph{scalable recipe} for navigation: one that minimizes sensor assumptions, generalizes across embodiments, and can be trained efficiently from simulation alone.

This scalability lens motivates every design decision in Robostral Navigate, an 8B vision-language model that achieves state-of-the-art performance on the Room-to-Room (R2R-CE)~\citep{r2r_ce} and Room-Across-Room (RxR-CE)~\citep{rxr} in Continuous Environments benchmarks using only a single RGB camera. A monocular RGB stream is the most universally available sensor across robotic platforms---from warehouse wheels to delivery drones---making it the natural input for a policy intended to scale. Rather than predicting metric displacements tied to a specific robot's geometry \citep{krantz2021waypoint, an2024etpnav}, Robostral Navigate produces waypoints via \emph{pointing}: it infers the image coordinates of the next target location in the robot's current camera view, together with the desired orientation upon arrival. Because waypoints are expressed in image space rather than metric coordinates, the policy avoids dependence on robot-specific geometry and transfers more naturally across camera configurations and embodiments.

The training recipe is designed with the same scalability objective. We train entirely in simulation~\citep{habitat19iccv, ramakrishnan2022hm3d} on approximately 2.4 million trajectories across 350k scenes, eliminating the need for costly real-world data collection. A prefix-caching technique based on attention masking compresses entire episodes into single training sequences, reducing training tokens by 22$\times$ while preserving all learning signals and transforming month-long training runs into days. After supervised training, we apply online reinforcement learning via CISPO~\citep{cispo}, training on a curated hard subset of trajectories to improve exploration and recovery from failures, yielding an additional 4\% improvement in success rate.

On R2R-CE validation unseen, Robostral Navigate achieves a 77.4\% success rate, surpassing the best single-camera method (Qwen-RobotNav-4B, 66.9\%) by 10.5 points and the best system using depth or multiple cameras (Qwen-RobotNav-8B, 72.1\%) by 5.3 points---despite relying on neither. Similarly, on the english-only RxR-CE validation unseen benchmark, it establishes a new state of the art with a 75.1\% success rate and a 68.7\% SPL, outperforming all single-camera baselines and demonstrating superior path efficiency over depth-assisted models, despite using only a single monocular RGB camera. This result demonstrates that a scalable, minimal-sensor recipe, when combined with efficient training and reinforcement learning, can outperform approaches that leverage privileged sensing.


\vspace{-0.025in}
\section{Modeling}
\label{sec:modeling}
\begin{figure}[ht]
\begin{center}
\includegraphics[width=0.99\linewidth,keepaspectratio]{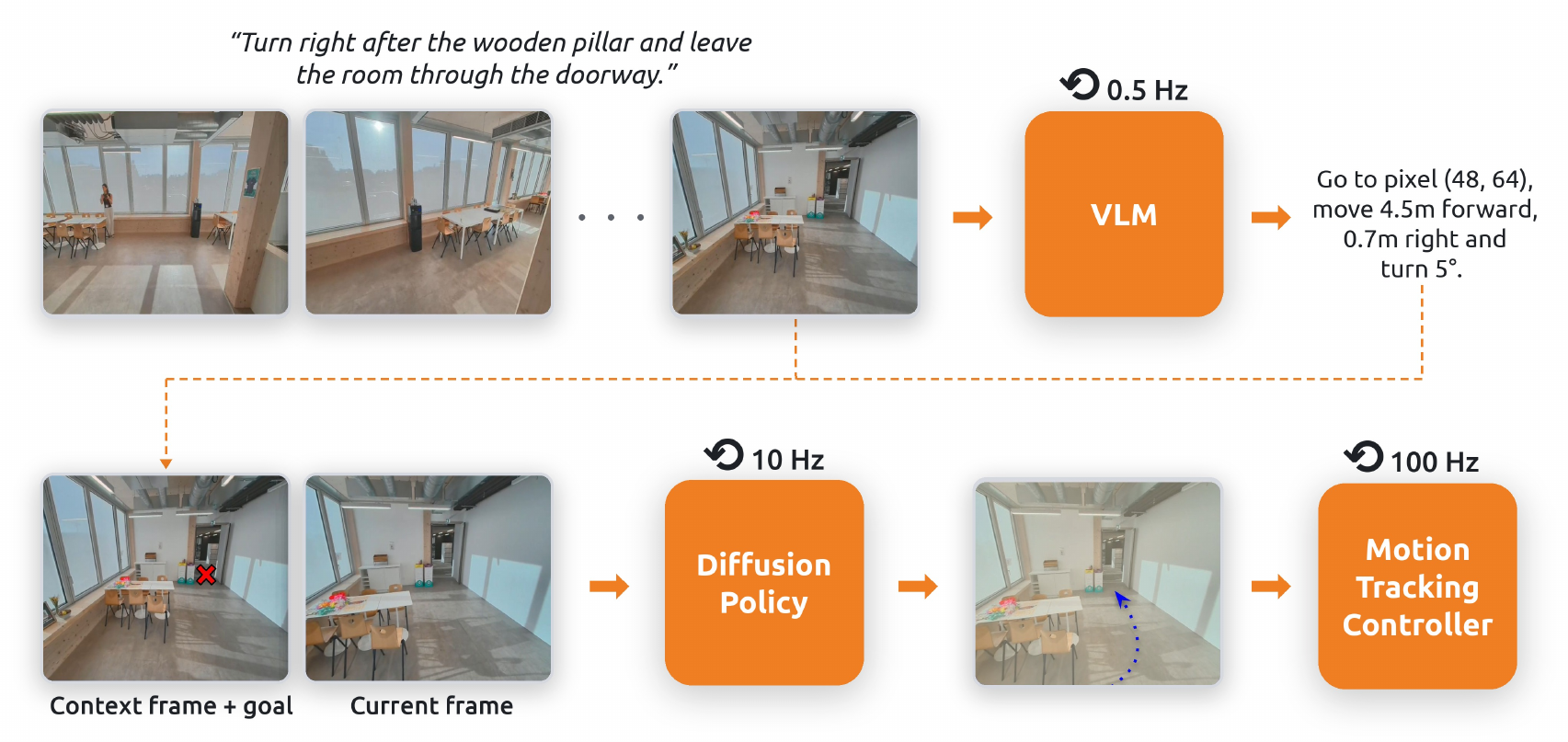}
\end{center}
\caption{\textbf{System Architecture.} Given a natural language instruction and a stream of RGB observations, the VLM predicts the next waypoint at 0.5 hertz via pointing, metric displacement and orientation upon arrival. When the target lies outside the field of view, pointing is omitted. A diffusion policy takes the waypoint, the context frame (frame at which the waypoint has been inferred) and the current frame to produce a dense trajectory at 10 hertz. Finally, a motion-tracking controller turns this trajectory into high-frequency actuator torques.}
\label{fig:architecture}
\end{figure}

In order to develop a scalable navigation system, we decompose the navigation task into two distinct challenges. First, the system must understand the navigation instruction, reason about the scene, and plan a coarse path for completing the task. Second, the coarse path must be translated into motor commands that efficiently move the robot towards the goal while avoiding obstacles. As illustrated in Figure~\ref{fig:architecture}, we built different models for each of these challenges. For the first, we trained a large 8B parameter vision-language model (VLM) called Robostral Navigate, which is needed for complex reasoning. For the second, we use a small 121M parameter diffusion policy, which has a sufficient capacity for the geometric understanding required for navigation. We combine these models in series and operate them at different inference frequencies.

\subsection{Architecture Overview}

Robostral Navigate takes a language instruction and history of monocular RGB frames as inputs, and outputs navigation waypoints in the form of pixel coordinates or displacement commands. The model is initialized from a dense 8B-parameter VLM designed for spatial grounding tasks such as pointing, counting, and object localization. Navigation is a natural extension of these grounding capabilities: once the model understands where things are in an image, it learns how to move toward them.

The frames are encoded by a vision encoder into a sequence of visual tokens, which are appended to the tokenized instruction to form the input sequence to the VLM. The history enables the model to reason about previously visited locations and track progress.

As illustrated in Figure~\ref{fig:architecture}, the inferred waypoints are converted to robot actions via a diffusion policy and a motion-tracking controller. Specifically, the diffusion policy outputs a low-level trajectory at 10 Hz in the form of an action chunk specifying the relative 2D displacements and 1D rotations for the next second. Then, an embodiement-specific motion-tracking controller converts the low-level trajectory into 100 Hz motor commands.

\subsection{Navigation via Pointing with Displacement Fallback}

Given a task instruction and a history of RGB frames, Robostral Navigate predicts the next destination using two modes: a preferred visual \emph{pointing} mode and a \emph{metric displacement} fallback.

In pointing mode, the model infers the image coordinates  $(u, v)$ of the furthest visible waypoint along the groundtruth trajectory, alongside the desired change of heading $\Delta\theta$ upon arrival, defined as the yaw rotation relative to the current frame.
Pointing is highly favored because it operates in visual space, making the policy naturally robust to changes in camera intrinsics and world scale, facilitating deployment across diverse robotics platforms.

However, pointing alone cannot handle cases where the destination lies outside the current field of view (e.g. when the robot should turn around).
To ensure the robot can always make progress, the system relies on a metric fallback mode consisting of local translational and rotational displacements $(\Delta x, \Delta y, \Delta\theta)$ relative to the current frame . 

In practice, we formulate training as a joint prediction task: when the destination is visible, the model predicts all five quantities
\begin{equation}
    a_{\text{vis}} = (u, v, \Delta x, \Delta y, \Delta\theta).
\end{equation}
Because only approximately 10\% of our training data features invisible destinations, predicting the metric displacements alongside pointing coordinates acts as an effective co-training task.
When no future position is visible from the current view, the model simply omits the image coordinates and only predicts metric displacements 
\begin{equation}
    a_{\text{invis}} = (\Delta x, \Delta y, \Delta\theta).
\end{equation}
The model also infers \text{STOP} when the task is completed.

\subsection{Pointing to Actions}

The Robotstral Navigate waypoints (specified via pointing or displacement) are grounded in the robot's camera frame. To convert these waypoints into robot actions, we use a lightweight diffusion transformer model~\citep{peebles2023scalable} with approximately 121M parameters. The inputs to the model are the VLM prediction $a \in \{a_{\text{vis}}, a_{\text{invis}}\}$, the robot's height and radius, the latest RGB frame used by the VLM $o_{t_{\text{vlm}}}$, and the current RGB frame $o_t$. Note that due to the VLM inference time, there is a latency between the context frame $o_{t_{\text{vlm}}}$ and current frame $o_t$. The output from the policy is an action chunk consisting of a sequence of 30 delta coordinates $(dx, dy, d\theta)$ relative to the robot base (same as the displacement waypoints) for the next second. Finally, a motion controller translates this action chunk into high-frequency motor commands.

\subsection{Cross-Robot Generalization}

\begin{figure}[htbp]
  \centering
  \includegraphics[width=0.65\linewidth,keepaspectratio]{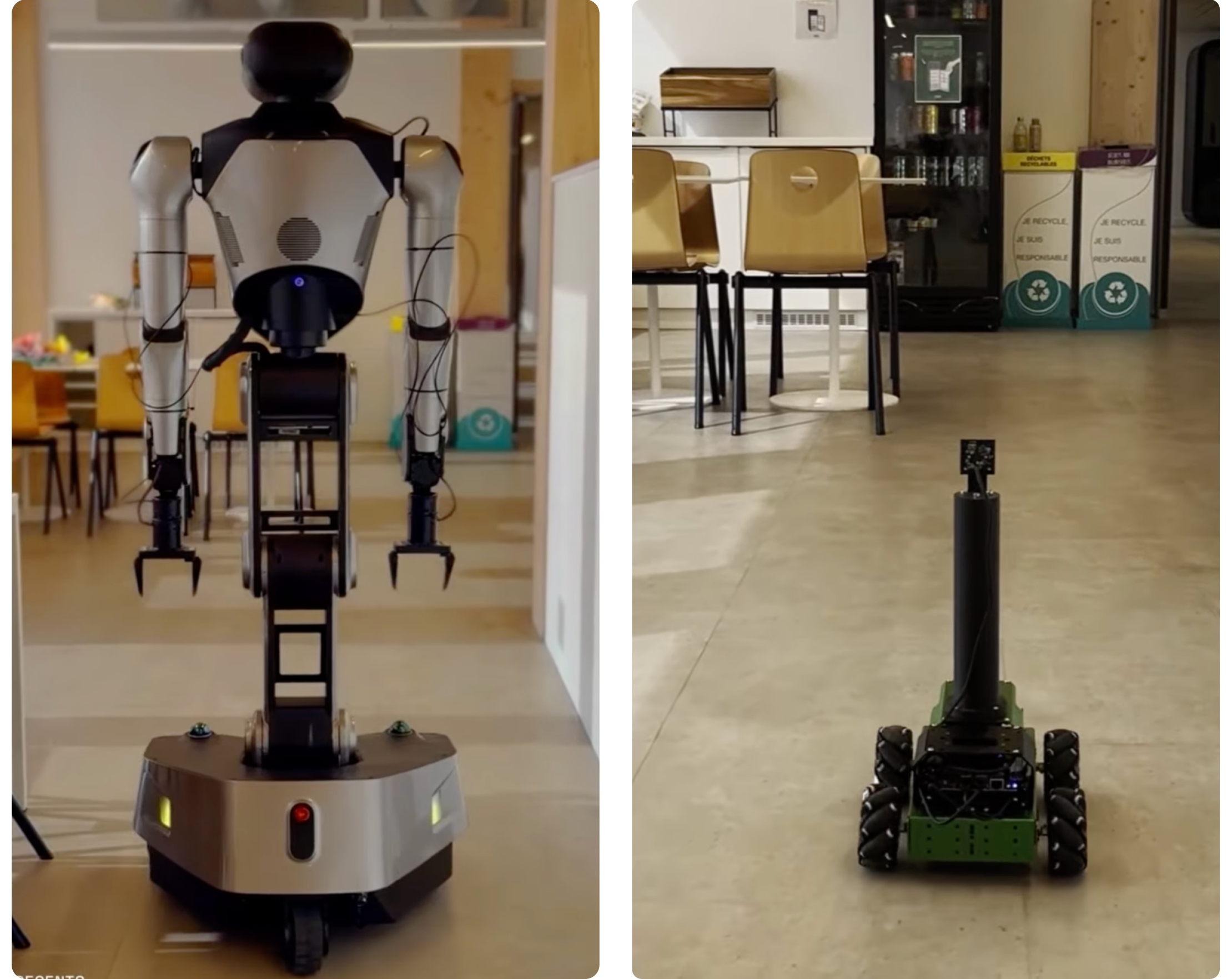}
  \caption{Cross-robot deployment of Robostral Navigate on the Galaxea R1 (left) and Hiwonder JetAuto (right). The two platforms differ substantially in robot morphology, camera height, camera configuration, and base geometry. Both deployments use the same Robostral Navigate vision-language model and diffusion policy weights, with only the low-level platform controller differing between robots.}
  \label{fig:cross_embodiment}
\end{figure}

Formulating navigation as a pointing task naturally enables cross-robot generalization. To further enhance cross-platform transfer, we randomize robot configurations when generating training data for both the VLM and diffusion policies. Specifically, we randomize the robot height between 0.4 - 1.8m, robot radius between 0.15 - 0.45m, camera placement height between 70 - 100\% of the robot height, and camera pitch between 0 - 25$^\circ$. These design choices reduce dependence on any particular camera setup, physical scale, or robot morphology, allowing the system to transfer across embodiments without retraining the learned policies. To demonstrate cross-robot generalization, we deploy the navigation system on two substantially different mobile robot platforms: the Galaxea R1~\citep{galaxeaR1} and the Hiwonder JetAuto~\citep{hiwonderJetAuto}. Notably, the same vision-language model and diffusion policy weights are used on both of the platforms illustrated in Figure~\ref{fig:cross_embodiment}.

\section{Training}
\label{sec:training}
\subsection{Data Generation}
\begin{figure}[H]
  \centering
  \includegraphics[width=0.99\linewidth,keepaspectratio]{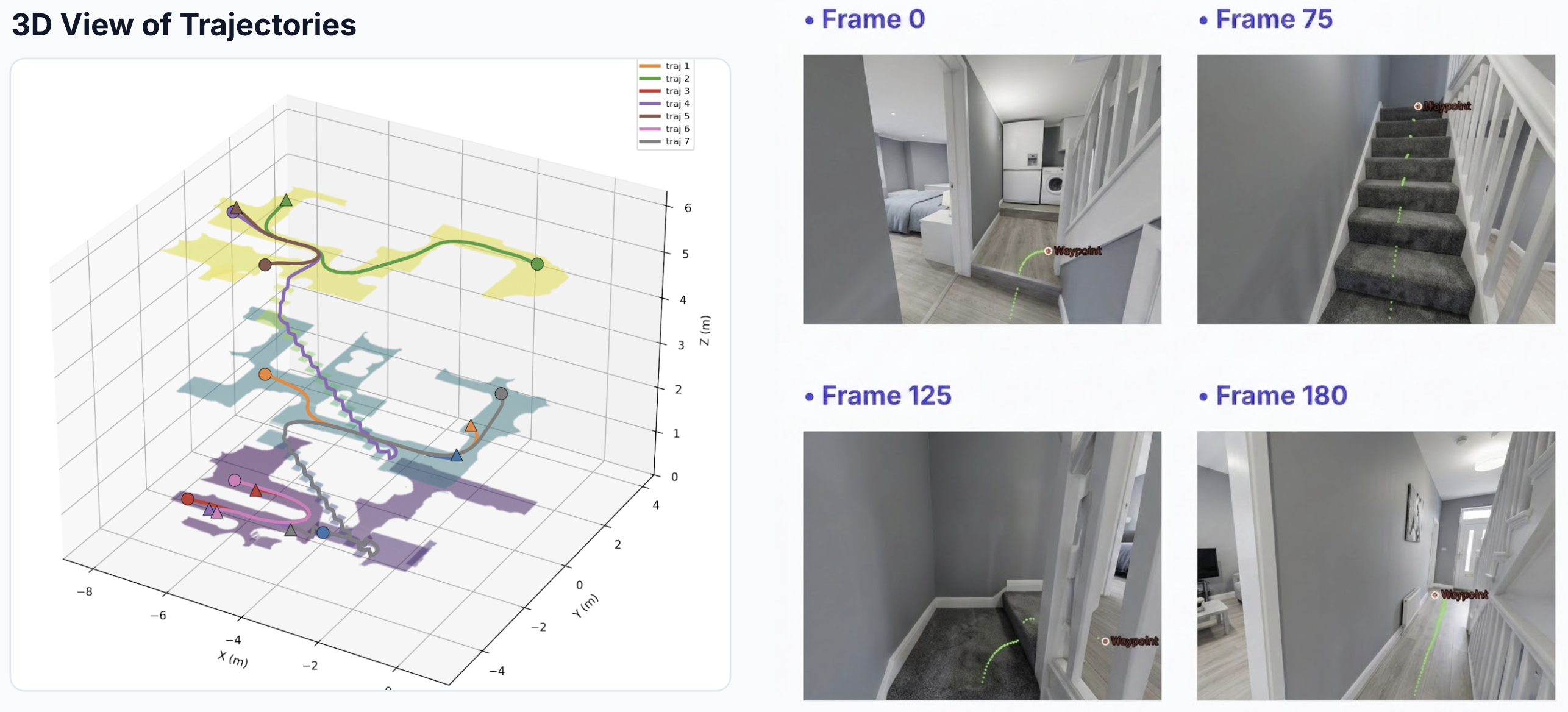}
    \caption{Example of training data. Left: 3D navigable area and sampled trajectories for a given scene. Right: Four frames from a cross-floor trajectory with future positions in-painted in green and waypoint in red.}
  \label{fig:data_fig}
\end{figure}

To enable rapid iteration and mitigate the reliance on costly real-world data collection, we built an efficient simulation-based data generation pipeline. 
The pipeline generates expert navigation trajectories across a diverse set of indoor and outdoor environments, producing approximately 2.4 million trajectories across 350k scenes. Each sample consists of a natural language instruction paired with a sequence of RGB observations and corresponding navigation actions.

As illustrated in Figure~\ref{fig:data_fig}, we use farthest point sampling to ensure diverse start and goal positions, resulting in trajectories of various length, sometimes spanning multiple floors. The diversity of scenes (offices, residential buildings, commercial spaces, and outdoors) is also a key factor for good generalization. We ensure variation in layout complexity, object density, lighting conditions, and architectural style.

\subsection{Supervised Training with Prefix Trees}

\begin{figure}[H]
  \centering
  \includegraphics[width=0.98\linewidth,keepaspectratio]{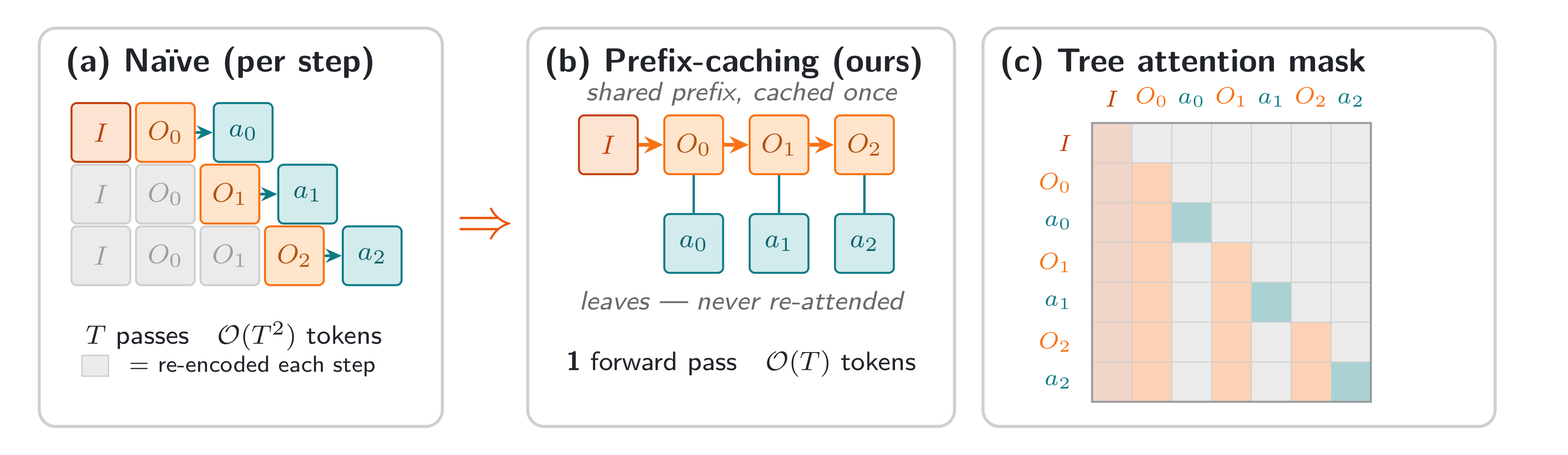}
    \caption{Comparison of (a) naive representation with our (b) prefix-caching. (c) demonstrates the tree attention mask with prefix-tree representation; I is the instruction, $O_i$ is the $i^\mathrm{th}$ observation/image, $a_i$ is the action at the $i^\mathrm{th}$ time step.}
  \label{fig:prefix_caching}
\end{figure}

A key ingredient of Robostral Navigate is an efficient training algorithm based on prefix sharing, shown in Figure~\ref{fig:prefix_caching}. Our method is inspired by the attention masking used in Molmo2~\citep{clark2026molmo2}, which amortizes the encoding of a context (text, image or video) shared across multiple independent questions and answers.
We push this concept further by exploiting the nested temporal structure of sequential decision-making: each iteration extends the history by concatenating the current observation, leading to significant redundancy between histories of the same episode.

In standard training, each time step is treated as an independent sample.
To predict the action at time step $t$, the model processes the instruction plus all the frames accumulated up to $t$.
A trajectory containing $T$ steps therefore produces $T$ separate training sequences, each containing an increasingly longer context. The total number of tokens processed scales as $\mathcal{O}(T^2)$.

\textbf{Efficiency.} Instead, we encode entire episodes as single training sequences consisting of an instruction followed by observations interleaved with actions $$I|O_0|a_0| \dots |O_n|a_n.$$
With this formulation, each new element in the history is encoded only once, allowing the losses for all time steps to be computed in a single forward pass and reducing the token complexity from $\mathcal{O}(T^2)$ to $\mathcal{O}(T)$.

However, with such sequences, vanilla causal attention would allow the model to attend the groundtruth actions associated with previous time steps during training. 
This is an issue because we do not have access to groundtruth actions during deployment and the model would heavily rely on these actions as previous groundtruth actions are often very informative of future actions (notably because of the farthest-visible waypoint paradigm).

\textbf{Attention masking.} To mitigate this issue, we introduce an attention masking strategy based on \textit{prefix trees}.
A prefix tree is a compact representation of a set of sequences, where shared prefixes are merged together in a trunk, while each sequence's specific suffix is encoded in its own branch.
For instance, if we have "AB" and "AC" then their flattened prefix-tree representation could be "ABC" where "A" would be the shared trunk, and "B" and "C" different branches.
The tree structure can be represented by an attention mask following the rule that  $token_i$ can attend to $token_j$ if (1) $j<i$ and (2) $token_j$ is in the trunk, or $token_i$ and $token_j$ are in the same branch.
A prefix tree sample gives provably identical training signal than the per-time-step samples used to build it, while saving compute by avoiding redundant prefix encoding during training.

This approach is particularly vital for the RxR-CE benchmark, which features significantly longer instructions and trajectory lengths compared to R2R-CE. Compared to per-time-step samples, our approach reduces the number of training tokens by $22\times$ without discarding any action-prediction targets. In practice, this reduces training time from months to days.

\subsection{Online Reinforcement Learning via CISPO}

\begin{figure}[!t]
  \centering
  \includegraphics[width=0.85\linewidth,keepaspectratio]{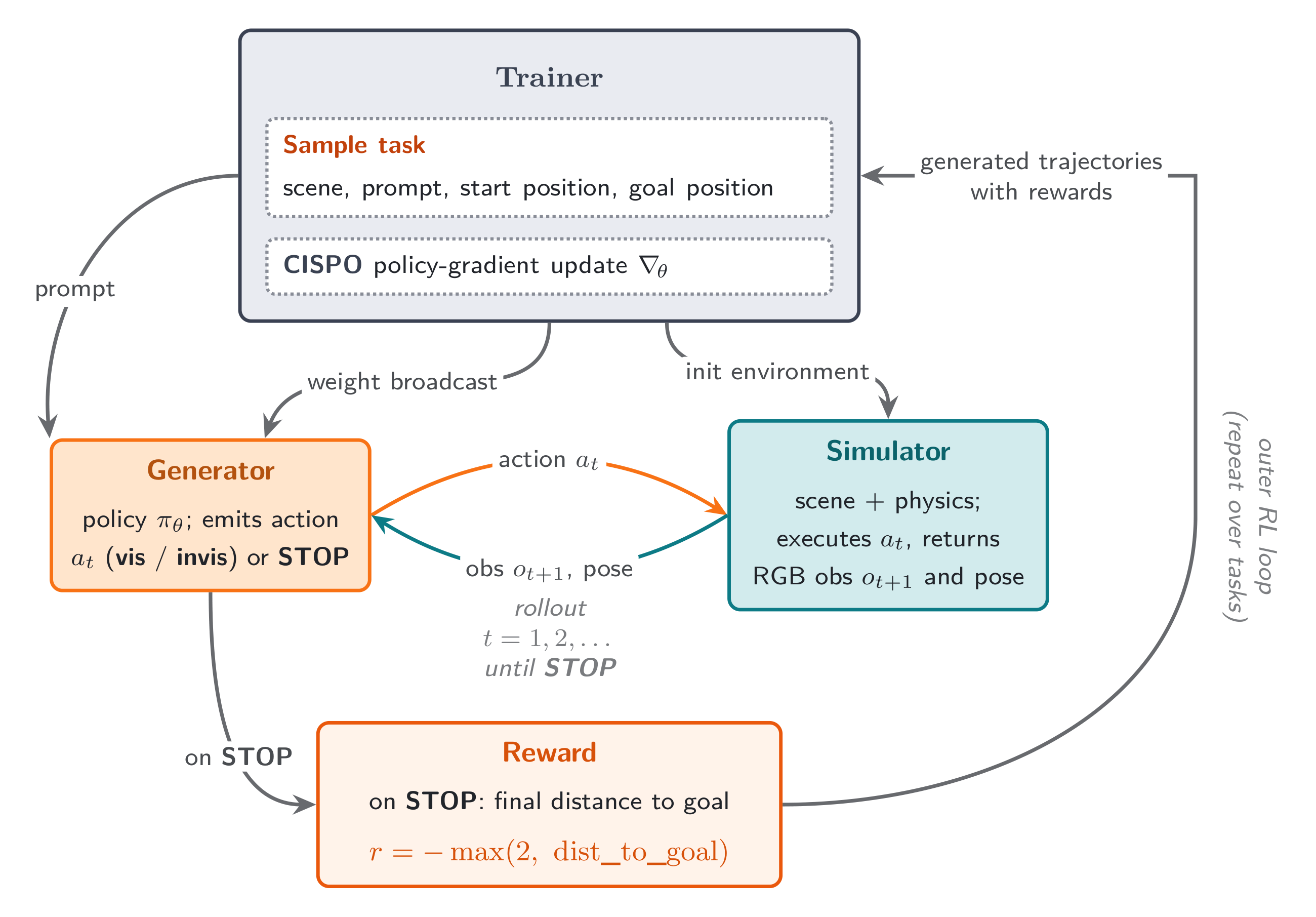}
  \caption{Overview of our online reinforcement learning pipeline. The trainer orchestrates the prompt sampling, physical simulator, and action generator (vLLM) in an asynchronous loop, updating the policy via CISPO using the $-{\max}(2, \text{dist\_to\_goal})$ reward signal upon episode termination.}
  \label{fig:rl_architecture}
\end{figure}

Although supervised fine-tuning (SFT) effectively instills a foundational visual-linguistic understanding and basic navigation skills, it fundamentally relies on behavioral cloning. This approach suffers from exposure bias~\citep{ranzato2016sequence}
and covariate shift~\citep{ross2011reduction}: at inference time,
compounding errors may lead the agent to out-of-distribution states. For instance, shortest-path trajectories do not demonstrate information-seeking and recovery behaviors, leading to poor performance when these abilities are needed.

To bridge this gap, we apply online reinforcement learning on top of the SFT model.
By actively interacting with a simulator, the agent can experience the consequences of its own actions and learn in the states it actually visits.
In this work, we use CISPO \citep{cispo}, leveraging group-relative advantage estimation to provide a stable, self-calibrated learning signal. 

The reward is defined as $-{\max}(2, \text{dist\_to\_goal})$, where $\text{dist\_to\_goal}$ is the geodesic distance between the last position and the goal in meters, derived with a path finding algorithm.
By clipping the distance penalty at 2 meters, the reward flatlines once the agent is near the goal location. This prevents the agent from needlessly maneuvering near the target to accumulate rewards that are unrelated to task success, and incentivizes the agent to emit the STOP action to end the episode.

Scaling this online RL pipeline for a large VLM is a formidable systems engineering challenge due to the intense hardware requirements. Unlike text-only RL setups, our architecture requires orchestrating three massive GPU-bound workloads concurrently: 1) the physics and rendering engine simulating hundreds of parallel environments, 2) the distributed inference engine (vLLM) generating autoregressive actions, and 3) the training nodes executing distributed weight updates.
Because the simulator itself consumes substantial GPU memory and compute to render high-fidelity observations, resource allocation had to be meticulously balanced across the cluster's nodes.
This careful orchestration ensures that the high-throughput vLLM generators and the training ranks do not stall while waiting for simulator rendering, maintaining optimal GPU utilization end-to-end.

\paragraph{Training tasks and visual curriculum.} We curate a \emph{hard subset} of 35k tasks by rolling out the SFT policy and retaining only the trajectories on which it fails to reliably reach the goal. This focuses compute on the cases where the policy most needs improvement -- complex layouts, ambiguous instructions, and long-horizon tasks, effectively filtering out episodes the model can already solve. During data collection, we pack tasks by scene, resulting in consecutive rollouts that are from the same scene. This induces an implicit visual curriculum: each batch contains multiple episodes sharing the same visual context. Empirically, this scene-contiguous ordering outperforms random shuffling.


\section{Results}
\label{sec:results}
We evaluate Robostral Navigate on the R2R-CE and RxR-CE benchmarks, which are the standard instruction-following benchmarks for navigation. Both require an agent to follow a previously unseen natural language instructions to navigate from a start to a goal position in previously unseen photorealistic environments. For these evaluations, we use a pathfinder from Habitat~\cite{habitat19iccv,szot2021habitat} for navigation between the waypoints predicted by Robostral Navigate.

\paragraph{Metrics:} We report standard navigation metrics that capture both task completion and trajectory efficiency.
\begin{itemize}[leftmargin=1.5em]
\item \textbf{Navigation Error (NE $\downarrow$)} measures the geodesic distance between the agent's final position and the goal. 
\item \textbf{Success Rate (SR $\uparrow$)} is the proportion of episodes where the agent stopped within 3\,m of the goal.
\item \textbf{Oracle Success (OS $\uparrow$)} is the proportion of episodes where the agent came within 3\,m of the goal at any point along its trajectory. 
\item  \textbf{Success weighted by Path Length (SPL $\uparrow$)} evaluates navigation efficiency by weighting SR using the ratio of the shortest-path length to the actual path length, penalizing unnecessary detours. 
\end{itemize}

\subsection{Main Results}

Table~\ref{tab:main_results} presents a comparison of Robostral Navigate against recent methods on R2R-CE and RxR-CE validation unseen. We categorize methods by whether they operate from a single RGB camera or require additional sensing modalities (depth, LiDAR, or multiple cameras).

\begin{table*}[t]
  \centering
  \caption{
  \textbf{Comparison on R2R-CE and RxR-CE validation unseen.} Robostral Navigate achieves state-of-the-art results using only a single RGB camera, outperforming methods that rely on depth sensors or multiple cameras on both benchmarks. \cmark{} indicates single-camera RGB-only; \xmark{} indicates use of depth, LiDAR, or multiple cameras.
  }
  \resizebox{\textwidth}{!}{
  \begin{tabular}{lcccccccc}
  \toprule
  \multirow{2}{*}{\textbf{Model}} & \multirow{2}{*}{\makecell{\textbf{Single}\\\textbf{Camera}}} & \multicolumn{4}{c}{\textbf{R2R-CE Val Unseen}} & \multicolumn{3}{c}{\textbf{RxR-CE Val Unseen}} \\
  \cmidrule(lr){3-6} \cmidrule(lr){7-9}
  & & \textbf{NE $\downarrow$} & \textbf{OS $\uparrow$} & \textbf{SR $\uparrow$} & \textbf{SPL $\uparrow$} & \textbf{NE $\downarrow$} & \textbf{SR $\uparrow$} & \textbf{SPL $\uparrow$} \\
  \midrule
  \multicolumn{9}{l}{\textit{Single RGB camera}} \\
  NaVid~\citep{navid} & \cmark & 5.47 & 0.490 & 0.370 & 0.350 & 5.72 & 0.457 & 0.382 \\
  Uni-NaVid~\citep{uni_navid} & \cmark & 5.58 & 0.535 & 0.470 & 0.427 & 6.24 & 0.487 & 0.409 \\
  NaVILA~\citep{navila} & \cmark & 5.22 & 0.625 & 0.540 & 0.490 & 6.77 & 0.493 & 0.440 \\
  InternVLA-N1~\citep{internvla_n1} & \cmark & 4.89 & 0.606 & 0.554 & 0.521 & 6.41 & 0.495 & 0.418 \\
  Qwen-VLA-Base~\citep{qwen_vla} & \cmark & 5.20 & 0.617 & 0.538 & 0.494 & 6.40 & 0.451 & 0.458 \\
  StreamVLN~\citep{streamvln} & \cmark & 4.98 & 0.642 & 0.569 & 0.519 & 6.22 & 0.529 & 0.460 \\
  Qwen-VLA-Instruct~\citep{qwen_vla} & \cmark & 5.10 & 0.690 & 0.575 & 0.512 & 5.80 & 0.596 & 0.478 \\
  Qwen-RobotNav-4B~\citep{qwen_robotnav} & \cmark & 4.22 & 0.736 & 0.669 & 0.605 & 4.15 & 0.713 & 0.615 \\
  Qwen-RobotNav-8B~\citep{qwen_robotnav} & \cmark & 4.36 & 0.727 & 0.657 & 0.596 & 4.16 & 0.734 & 0.635 \\
  \midrule
  \multicolumn{9}{l}{\textit{Depth / multi-camera}} \\
  InternVLA-N1 (S1+S2)~\citep{internvla_n1} & \xmark & 4.83 & 0.633 & 0.582 & 0.540 & 5.91 & 0.535 & 0.461 \\
  NavFoM~\citep{navfom} & \xmark & 4.61 & 0.721 & 0.617 & 0.553 & 4.74 & 0.644 & 0.562 \\
  ABot-N0~\citep{abot_n0} & \xmark & 3.78 & 0.708 & 0.664 & 0.639 & 3.83 & 0.693 & 0.600 \\
  OmniNav~\citep{omninav} & \xmark & 3.74 & 0.746 & 0.695 & 0.661 & 3.77 & 0.736 & 0.620 \\
  Qwen-RobotNav-4B~\citep{qwen_robotnav} & \xmark & 3.80 & 0.772 & 0.695 & 0.636 & 3.80 & 0.752 & 0.650 \\
  Qwen-RobotNav-8B~\citep{qwen_robotnav} & \xmark & 3.73 & 0.727 & 0.721 & 0.666 & 3.58 & \textbf{0.765} & 0.657 \\
  \midrule
  \textbf{Robostral Navigate (ours)} & \cmark & \textbf{3.20} & \textbf{0.813} & \textbf{0.774} & \textbf{0.742} & \textbf{3.47} & 0.751 & \textbf{0.687} \\
  \bottomrule
  \end{tabular}
  }
  \label{tab:main_results}
\end{table*}

Robostral Navigate achieves a success rate of 77.4\% and SPL of 74.2\% on R2R-CE validation unseen, establishing a new state of the art across all metrics. It surpasses the best prior single-camera method (Qwen-RobotNav-4B at 66.9\% SR) by 10.5 points, and the best method using depth or multiple cameras (Qwen-RobotNav-8B at 72.1\% SR) by 5.3 points---despite using only a single RGB camera with no auxiliary sensors.

In the R2R-CE validation seen split, Robostral Navigate achieves a 79.4\% success rate, demonstrating strong performance in both seen and unseen environments.
This performance is directly unlocked by our online RL phase, which profoundly shifts the policy from brittle to robust navigation. Compared to our R2R-CE SFT baseline (75.96\% seen, 73.40\% unseen), our post-RL checkpoint achieves a success rate of 79.43\% on seen environments (+3.47\%) and 77.43\% on unseen environments (+4.03\%). Peak validation performance during the run reached 80.46\% on seen and 77.43\% on unseen environments, highlighting the robust generalization capabilities unlocked by our online RL pipeline. The high SPL (74.2\%) shows that these successes are achieved via efficient paths, without unnecessary exploration.

On the more challenging RxR-CE benchmark, Robostral Navigate continues to demonstrate superior performance. It achieves a success rate of 75.1\% and an SPL of 68.7\%, while maintaining a low navigation error of only 3.47 meters. Compared to the prior state-of-the-art single-camera method, Qwen-RobotNav-8B (73.4\% SR, 63.5\% SPL), Robostral Navigate improves the success rate by 1.7 percentage points and significantly boosts path efficiency (SPL) by 5.2 points. Remarkably, our single-camera, RGB-only model even outperforms the best depth- and multi-camera-assisted baseline (Qwen-RobotNav-8B with depth) on both SPL (68.7\% vs. 65.7\%) and navigation error (3.47m vs. 3.58m), while remaining competitive on success rate (75.1\% vs. 76.5\%).

Similar to the trends on R2R-CE, on the RxR-CE benchmark, the online RL phase brings a substantial boost over the SFT baseline.
Robostral Navigate improves from 71.2\% to a state-of-the-art success rate of 75.1\% (+3.9\%), demonstrating that reinforcement learning generalizes well to long-horizon instruction-following tasks.
These results underscore Robostral Navigate's exceptional ability to handle long, complex instructions and plan highly optimal trajectories without relying on auxiliary depth or panoramic sensors.




\section{Conclusion}
\label{sec:conclusion}

We introduced Robostral Navigate, an 8B vision-language model for embodied navigation that achieves state-of-the-art performance on the R2R-CE and RxR-CE benchmarks. Robostral Navigate is designed with scalability in mind to minimize sensor assumptions, generalize across embodiments, and train efficiently from simulation alone. The policy predicts navigation actions via pointing and only needs a single RGB camera to be deployed. The policy is trained with a prefix-caching technique that cuts training tokens by 22$\times$ while preserving all learning signals, resulting in a highly efficient supervised finetuning training recipe. Finally, the policy is finetuned using reinforcement learning with CISPO, resulting in an agent that is very efficient at solving navigation tasks (as measured on the R2R-CE and RxR-CE benchmarks).

Navigation is a foundational capability for general-purpose robotics. Accordingly, we view Robostral Navigate as a first step towards building a general-purpose embodied agent. Specifically, this work demonstrates that combining large-scale simulation, efficient training, and reinforcement learning can produce a state-of-the-art embodied navigation model that has minimal sensing requirements.


\subsection*{Core contributors}

Arjun Majumdar,
Avinash Sooriyarachchi,
Benjamin Tibi,
Chris Bamford,
Elliot Chane-Sane,
Guillaume Lample,
Khyathi Raghavi Chandu,
Ludovic Ho Fuh,
Mathieu Poir\'{e}e,
Olivier Duchenne,
Rosalie Millner,
Srijan Mishra,
Th\'{e}o Cachet,
Thomas Chabal.

\subsection*{Contributors}
Abdelaziz Bounhar,
Abhijeet Somani,
Aditi Kabra,
Adrian Valente,
Adrien Petralia,
Adrien Sad\'e,
Alan Jeffares,
Albert Jiang,
Aleksandr Timashov,
Alexandre Cahill,
Alexandre Gavaudan,
Alexandre Laval,
Alexandre Sablayrolles,
Am\'elie H\'eliou,
Amos You,
Andr\'e Jonasson,
Andrew Bai,
Andrew Ehrenberg,
Andrew Zhao,
Angele Lenglemetz,
Anmol Agarwal,
Antonia Calvi,
Arata Suzuki,
Arthur Fournier,
Artjom Joosen,
Aylin Guliz Akkus,
Aysenur Karaduman,
Baptiste Bout,
Baptiste Rozi\`ere,
Baudouin De Monicault,
Benjamin Holzschuh,
Benjamin Lefaudeux,
Bernhard Stadlbauer,
B\l{}a\.zej Osi\'nski,
Camille Le Scao,
Chaoran Yu,
Charlotte Cronj\"ager,
Chen-Yo Sun,
Christian Wallenwein,
Christophe Renaudin,
Cl\'emence Lanfranchi,
Corentin Barreau,
Corentin Sautier,
Cristiana-Diana Diaconu,
Cyprien Courtot,
Daniel Marczak,
Darius Dabert,
Diego de Las Casas,
Dominik Nuss,
Dylan Rubini,
Dzmitry Soupel,
Elizaveta Demyanenko,
Emilien Fugier,
Emmanuel Gottlob,
Erik Aas,
\'Etienne Millon,
Etienne Goffinet,
Eujeong Choi,
Fabian Paischer,
Fabian Schlager,
Faruk Ahmed,
Federico Baldassarre,
Filip Szatkowski,
Florian Wiesner,
Gabrielle Berrada,
Ga\"etan Ecrepont,
Ga\'etan Lepage,
Gaspard Blanchet,
Gaspard Donada-Vidal,
Gauthier Delerce,
Gauthier Guinet,
Genevieve Hayes,
Georgii Novikov,
Giada Pistilli,
Gianluca Galletti,
Guillaume Breton,
Guillaume Kunsch,
Guillaume Martin,
Guillaume Raille,
Gunjan Dhanuka,
Gunshi Gupta,
Han Zhou,
Harshil Shah,
Hasan Furkan Vural,
H\'edi Hadiji,
Hope McGovern,
Hugo Cisneros,
Hugo Thimonier,
Indraneel Mukherjee,
Ivan Cuevas Salazar,
Jacques Sun,
Jan Ludziejewski,
Jason Rute,
Jean Quentin,
Jean-Hadrien Chabran,
Jean-Malo Delignon,
Jie Zhang,
Joachim Studnia,
Joep Barmentlo,
Johannes Brandstetter,
John Harvill,
Jonas Amar,
Jonas Schweizer,
Jos\'ephine Delas,
Josselin Somerville,
Julien Denize,
Julien Tauran,
Kartik Khandelwal,
Kilian Tep,
Kush Jain,
Larissa Laich,
Laura Calem,
Laurence Aitchison,
Laurent Callot,
Laurent Fainsin,
L\'eo Cotteleer,
L\'eonard Blier,
Lingxiao Zhao,
Louis Martin,
Louis Serrano,
Lucile Saulnier,
Luis Montero,
Maarten Buyl,
Manon Chossegros,
Marcin Mo\.zejko,
Margaret Jennings,
Markus Hennerbichler,
Martin Alexandre,
Mathieu Schmitt,
Mathilde Guillaumin,
Matthieu Andr\'e,
Matthieu Dinot,
Matthieu Futeral,
Maurits Bleeker,
Mauro Comi,
Max Mynter,
Maxim Berman,
Maxime Darrin,
Maxime Louis,
Maximilian Augustin,
Maximilian M\"uller,
Melina Jingting Laimon,
Mert Unsal,
Mia Chiquier,
Michael Pilcer,
Micha\l{} Pietruszka,
Micha\l{} Zaj\k{a}c,
Mikhail Biriuchinskii,
Minh-Quang Pham,
Minwoo Kang,
Morgane Rivi\`ere,
Namit Katariya,
Nathan Grinsztajn,
Nathan Simpson,
Neeraj Aggarwal,
Neha Gupta,
Ola Mysiak,
Oliver Leicht,
Olivier Bousquet,
Parag Jain,
Patricia Wang,
Patrick Blies,
Patrick von Platen,
Paul Jacob,
Paul Wambergue,
Paula Kurylowicz,
Pavan Kumar Reddy,
Pavel Kuksa,
Philippe Pinel,
Philom\`ene Chagniot,
Pierre Stock,
Pierre-Andr\'e Savalle,
Piotr Milos,
Prateek Gupta,
Pravesh Agrawal,
Quentin Desreumaux,
Quentin Torroba,
Quercus Hernandez,
Ram Ramrakhya,
Randall Isenhour,
Ranjit Parva,
Raul Perez Pelaez,
Reinhard Sonnleitner,
R\'emi Delacourt,
Richard Kurle,
Rishi Shah,
Rob Romijnders,
Rohin Arora,
Romain Sauvestre,
Roman Soletskyi,
Rupert Menneer,
Sagar Vaze,
Samuel Barry,
Samuel Belkadi,
Samuel Humeau,
Sanchit Gandhi,
Sandeep Subramanian,
Sarthak Mittal,
Saskia Adaime,
Sean Cha,
Sebastian Kaltenbach,
Shashwat Dalal,
Shashwat Verma,
Sherif Waly,
Shrimai Prabhumoye,
Siddhant Waghjale,
Siddharth Gandhi,
Simon Lepage,
Simon Sorg,
Soham Ghosh,
Sophie Marbach,
Stanislas Lange,
Steve Hong,
Sumukh Aithal,
Szymon Antoniak,
Tarun Kumar Vangani,
Teven Le Scao,
Thibaut Lavril,
Thomas Coste,
Thomas Defard,
Thomas Foubert,
Thomas Robert,
Thomas Wang,
Tianyu Zhang,
Tim Lawson,
Timoth\'ee Lacroix,
Tobias Kronlachner,
Tom Bewley,
Tom Edwards,
Tomas Hodan,
Tuhin Das,
Tyler Wang,
Ulrick BLE,
Umar Jamil,
Umberto Tomasini,
Valentin Mac\'e,
Van Phung,
Vedant Nanda,
Victor Jouault,
Victor Letzelter,
Victor Paltz,
Victor Poucheret,
Vincent Maladi\`ere,
Vincent Pfister,
Virgile Richard,
Vladislav Bataev,
Wassim Bouaziz,
Wen Ding Li,
William Havard,
William Marshall,
Xinghui Li,
Xingran Guo,
Xinyu Yang,
Yann Dreze,
Yassine El Ouahidi,
Yassir Bendou,
Yihan Wang,
Yimu Pan,
Yves Martin des Taillades,
Zaccharie Ramzi,
Zhenlin Xu,
Zsofia Csakany.

\clearpage


\bibliography{ref}

@inproceedings{navid,
  title     = {{NaVid}: Video-based {VLM} Plans the Next Step for Vision-and-Language Navigation},
  author    = {Zhang, Jiazhao and Wang, Kunyu and Xu, Rongtao and Zhou, Gengze and Hong, Yicong and Fang, Xiaomeng and Wu, Qi and Zhang, Zhizheng and Wang, He},
  booktitle = {Robotics: Science and Systems},
  year      = {2024},
  url       = {https://arxiv.org/abs/2402.15852}
}

@article{uni_navid,
  title   = {{Uni-NaVid}: A Video-based Vision-Language-Action Model for Unifying Embodied Navigation Tasks},
  author  = {Zhang, Jiazhao and Wang, Kunyu and Wang, Shaoan and Li, Minghan and Liu, Haoran and Wei, Songlin and Wang, Zhongyuan and Zhang, Zhizheng and Wang, He},
  journal = {arXiv preprint arXiv:2412.06224},
  year    = {2024}
}

@inproceedings{navila,
  title     = {{NaVILA}: Legged Robot Vision-Language-Action Model for Navigation},
  author    = {Cheng, An-Chieh and Ji, Yandong and Yang, Zhaojing and Gongye, Zaitian and Zou, Xueyan and Kautz, Jan and B{\i}y{\i}k, Erdem and Yin, Hongxu and Liu, Sifei and Wang, Xiaolong},
  booktitle = {Robotics: Science and Systems},
  year      = {2025},
  url       = {https://arxiv.org/abs/2412.04453}
}

@article{internvla_n1,
  title   = {{Ground Slow, Move Fast}: A Dual-System Foundation Model for Generalizable Vision-and-Language Navigation},
  author  = {Jiazhao Zhang and Anqi Li and Yunpeng Qi and Minghan Li and Jiahang Liu and Shaoan Wang and Haoran Liu and Gengze Zhou and Yuze Wu and Xingxing Li and Yuxin Fan and Wenjun Li and Zhibo Chen and Fei Gao and Qi Wu and Zhizheng Zhang and He Wang},
  journal = {arXiv preprint arXiv:2512.08186},
  year    = {2025}
}

@article{streamvln,
  title   = {{StreamVLN}: Streaming Vision-and-Language Navigation via SlowFast Context Modeling},
  author  = {Meng Wei and Chenyang Wan and Xiqian Yu and Tai Wang and Yuqiang Yang and Xiaohan Mao and Chenming Zhu and Wenzhe Cai and Hanqing Wang and Yilun Chen and Xihui Liu and Jiangmiao Pang},
  journal = {arXiv preprint arXiv:2507.05240},
  year    = {2025}
}

@article{qwen_vla,
  title   = {{Qwen-VLA}: Unifying Vision-Language-Action Modeling across Tasks, Environments, and Robot Embodiments},
  author  = {Qiuyue Wang and others},
  journal = {arXiv preprint arXiv:2605.30280},
  year    = {2026}
}

@article{qwen_robotnav,
  title   = {{Qwen-RobotNav Technical Report}: A Scalable Navigation Model Designed for an Agentic Navigation System},
  author  = {Zhang, Jiazhao and Zhou, Gengze and Yin, Hale and Huang, Yiyang and Lei, Zixing and Peng, Qihang and Yuan, Haoqi and Zhang, Jie and Guo, Xudong and Chen, Xiaoyue and others},
  journal = {arXiv preprint arXiv:2606.18112},
  year    = {2026}
}

@article{navfom,
  title   = {Embodied Navigation Foundation Model},
  author  = {Jiazhao Zhang and Anqi Li and Yunpeng Qi and Minghan Li and Jiahang Liu and Shaoan Wang and Haoran Liu and Gengze Zhou and Yuze Wu and Xingxing Li and Yuxin Fan and Wenjun Li and Zhibo Chen and Fei Gao and Qi Wu and Zhizheng Zhang and He Wang},
  journal = {arXiv preprint arXiv:2509.12129},
  year    = {2025}
}

@article{abot_n0,
  title   = {{ABot-N0}: Technical Report on the {VLA} Foundation Model for Versatile Embodied Navigation},
  author  = {Chu, Zedong and Xie, Shichao and Wu, Xiaolong and Shen, Yanfen and Luo, Minghua and Wang, Zhengbo and Liu, Fei and Leng, Xiaoxu and Hu, Junjun and Yin, Mingyang and others},
  journal = {arXiv preprint arXiv:2602.11598},
  year    = {2026}
}

@inproceedings{omninav,
  title     = {{OmniNav}: A Unified Framework for Prospective Exploration and Visual-Language Navigation},
  author    = {Xinda Xue and Junjun Hu and Minghua Luo and Shichao Xie and Jintao Chen and Zixun Xie and Kuichen Quan and Wei Guo and Mu Xu and Zedong Chu},
  booktitle = {International Conference on Learning Representations},
  year      = {2026},
  url       = {https://arxiv.org/abs/2509.25687}
}

@inproceedings{r2r_ce,
  title     = {Beyond the {Nav-Graph}: Vision-and-Language Navigation in Continuous Environments},
  author    = {Jacob Krantz and Erik Wijmans and Arjun Majumdar and Dhruv Batra and Stefan Lee},
  booktitle = {European Conference on Computer Vision},
  year      = {2020},
  url       = {https://arxiv.org/abs/2004.02857}
}

@article{cispo,
  title   = {{MiniMax-M1}: Scaling Test-Time Compute Efficiently with Lightning Attention},
  author  = {Chen, Aili and Li, Aonian and Gong, Bangwei and Jiang, Binyang and Fei, Bo and Yang, Bo and Shan, Boji and Yu, Changqing and Wang, Chao and Zhu, Cheng and others},
  journal = {arXiv preprint arXiv:2506.13585},
  year    = {2025}
}

@inproceedings{ross2011reduction,
  title     = {A reduction of imitation learning and structured prediction to no-regret online learning},
  author    = {Ross, St{\'e}phane and Gordon, Geoffrey and Bagnell, Drew},
  booktitle = {Proceedings of the fourteenth international conference on artificial intelligence and statistics},
  year      = {2011},
  organization = {JMLR Workshop and Conference Proceedings}
}

@inproceedings{peebles2023scalable,
  title={Scalable diffusion models with transformers},
  author={Peebles, William and Xie, Saining},
  booktitle={International Conference on Computer Vision},
  pages={4195--4205},
  year={2023}
}

@inproceedings{habitat19iccv,
  title     = {Habitat: A Platform for Embodied {AI} Research},
  author    = {Manolis Savva and Abhishek Kadian and Oleksandr Maksymets and Yili Zhao and Erik Wijmans and Bhavana Jain and Julian Straub and Jia Liu and Vladlen Koltun and Jitendra Malik and Devi Parikh and Dhruv Batra},
  booktitle = {International Conference on Computer Vision},
  year      = {2019},
  url       = {https://arxiv.org/abs/1904.01201}
}

@inproceedings{szot2021habitat,
  title     =     {Habitat 2.0: Training Home Assistants to Rearrange their Habitat},
  author    =     {Andrew Szot and Alex Clegg and Eric Undersander and Erik Wijmans and Yili Zhao and John Turner and Noah Maestre and Mustafa Mukadam and Devendra Chaplot and Oleksandr Maksymets and Aaron Gokaslan and Vladimir Vondrus and Sameer Dharur and Franziska Meier and Wojciech Galuba and Angel Chang and Zsolt Kira and Vladlen Koltun and Jitendra Malik and Manolis Savva and Dhruv Batra},
  booktitle =     {Advances in Neural Information Processing Systems (NeurIPS)},
  year      =     {2021}
}

@inproceedings{shah2023lm,
  title     = {{LM-Nav}: Robotic Navigation with Large Pre-Trained Models of Language, Vision, and Action},
  author    = {Dhruv Shah and Blazej Osinski and Brian Ichter and Sergey Levine},
  booktitle = {Conference on Robot Learning},
  year      = {2023},
  url       = {https://arxiv.org/abs/2207.04429}
}

@inproceedings{clark2026molmo2,
  title={Molmo2: Open weights and data for vision-language models with video understanding and grounding},
  author={Clark, Christopher and Zhang, Jieyu and Ma, Zixian and Park, Jae Sung and Salehi, Mohammadreza and Tripathi, Rohun and Lee, Sangho and Ren, Zhongzheng and Kim, Chris Dongjoo and Yang, Yinuo and Shao, Vincent and Yang, Yue and Huang, Weikai and Gao, Ziqi and Anderson, Taira and Zhang, Jianrui and Jain, Jitesh and Stoica, George and Han, Winson and Farhadi, Ali and Krishna, Ranjay},
  booktitle={Conference on Computer Vision and Pattern Recognition},
  year={2026}
}

@inproceedings{ranzato2016sequence,
  title     = {Sequence Level Training with Recurrent Neural Networks},
  author    = {Ranzato, Marc'Aurelio and Chopra, Sumit and Auli, Michael and Zaremba, Wojciech},
  booktitle = {International Conference on Learning Representations},
  year      = {2016}
}

@misc{galaxeaR1,
  author       = {{Galaxea AI}},
  title        = {Galaxea R1},
  howpublished = {\url{https://userguide-galaxea.github.io/Product_User_Guide/Introducing_Galaxea_Robot/product_info/R1/}},
  note         = {Accessed: 2026-07-15},
  year         = {2026}
}

@misc{hiwonderJetAuto,
  author       = {{Hiwonder}},
  title        = {JetAuto ROS Robot},
  howpublished = {\url{https://developer.nvidia.com/embedded/community/jetson-projects/jetauto}},
  note         = {Accessed: 2026-07-15},
  year         = {2026}
}

@inproceedings{brohan2023rt2,
  title={{RT-2}: Vision-language-action models transfer web knowledge to robotic control},
  author={Zitkovich, Brianna and Yu, Tianhe and Xu, Sichun and Xu, Peng and Xiao, Ted and Xia, Fei and Wu, Jialin and Wohlhart, Paul and Welker, Stefan and Wahid, Ayzaan and others},
  booktitle={Conference on Robot Learning},
  year={2023},
  organization={PMLR}
}

@inproceedings{krantz2021waypoint,
  title     = {Waypoint Models for Instruction-guided Navigation in Continuous Environments},
  author    = {Jacob Krantz and Aaron Gokaslan and Dhruv Batra and Stefan Lee and Oleksandr Maksymets},
  booktitle = {International Conference on Computer Vision},
  year      = {2021},
  url       = {https://arxiv.org/abs/2110.02207}
}

@article{an2024etpnav,
  title   = {{ETPNav}: Evolving Topological Planning for Vision-Language Navigation in Continuous Environments},
  author={An, Dong and Wang, Hanqing and Wang, Wenguan and Wang, Zun and Huang, Yan and He, Keji and Wang, Liang},
  journal = {IEEE Transactions on Pattern Analysis and Machine Intelligence},
  year    = {2024},
  publisher={IEEE}
}

@inproceedings{ramakrishnan2022hm3d,
  title     = {Habitat-Matterport {3D} Dataset ({HM3D}): 1000 Large-scale {3D} Environments for Embodied {AI}},
  author    = {Santhosh Kumar Ramakrishnan and Aaron Gokaslan and Erik Wijmans and Oleksandr Maksymets and Alexander Clegg and John Turner and Eric Undersander and Wojciech Galuba and Andrew Westbury and Angel X. Chang and Manolis Savva and Yili Zhao and Dhruv Batra},
  booktitle = {Neural Information Processing Systems Datasets and Benchmarks Track},
  year      = {2022},
}

@inproceedings{rxr,
  title={{Room-across-Room}: Multilingual vision-and-language navigation with dense spatiotemporal grounding},
  author={Ku, Alexander and Anderson, Peter and Patel, Roma and Ie, Eugene and Baldridge, Jason},
  booktitle={Conference on Empirical Methods in Natural Language Processing},
  pages={4392--4412},
  year={2020}
}

\end{document}